\documentclass[letterpaper, 10 pt, conference]{ieeeconf}  

\IEEEoverridecommandlockouts                              

\usepackage{amsmath}
\usepackage{graphicx}
\usepackage{array} 
\usepackage{booktabs}
\usepackage{upgreek}
\usepackage{multirow} 
\usepackage{array} 
\usepackage{amssymb}
\usepackage{caption}

\title{\LARGE \bf
Fake It To Make It: Virtual Multiviews to Enhance \\ Monocular Indoor Semantic Scene Completion
}

\author{Anith Selvakumar and Manasa Bharadwaj$^{1}$
\thanks{$^{1}$Anith Selvakumar and Manasa Bharadwaj are with LG Electronics, Toronto AI Lab, Canada.
    {\tt\small a.selvakumarasingam@lge.com}, {\tt\small manasa.bharadwaj@lge.com}
    }%
}

\begin{document}

\maketitle
\thispagestyle{empty}
\pagestyle{empty}

\begin{abstract}

Monocular Indoor Semantic Scene Completion (SSC) aims to reconstruct a 3D semantic occupancy map from a single RGB image of an indoor scene, inferring spatial layout and object categories from 2D image cues. The challenge of this task arises from the depth, scale, and shape ambiguities that emerge when transforming a 2D image into 3D space, particularly within the complex and often heavily occluded environments of indoor scenes. Current SSC methods often struggle with these ambiguities, resulting in distorted or missing object representations. To overcome these limitations, we introduce an innovative approach that leverages novel view synthesis and multiview fusion. Specifically, we demonstrate how virtual cameras can be placed around the scene to emulate multiview inputs that enhance contextual scene information. We also introduce a Multiview Fusion Adaptor (MVFA) to effectively combine the multiview 3D scene predictions into a unified 3D semantic occupancy map. Finally, we identify and study the inherent limitation of generative techniques when applied to SSC, specifically the Novelty-Consistency tradeoff. Our system, GenFuSE, demonstrates IoU score improvements of up to 2.8\% for Scene Completion and 4.9\% for Semantic Scene Completion when integrated with existing SSC networks on the NYUv2 dataset. This work introduces GenFuSE as a standard framework for advancing monocular SSC with synthesized inputs.

\end{abstract}

\section{INTRODUCTION}


Robust perception modules, employing computer vision tasks such as object detection, semantic segmentation, and occupancy prediction, are fundamental to embodied AI systems. These modules enable agents to learn and interact with their environments across a diverse set applications, including autonomous vehicles (AV), automated manufacturing, and robotic systems \cite{Xu2025}\cite{Li2024}\cite{Liu2023}\cite{wang2023embodiedscan}.

Semantic Scene Completion (SSC) exemplifies a critical perception module that integrates semantic segmentation and occupancy prediction into a comprehensive scene understanding task \cite{Song2017}. SSC aims to generate a complete 3D representation of a scene, including both geometric structure (occupancy) and semantic labels (object categories) for each point in 3D space, from a single RGB or RGB-D image. The inherent challenge of reconstructing a 3D representation from a 2D image stems from the lack of information of occluded objects or regions outside the camera's view. Consequently, SSC models must learn object shapes, spatial relationships, and general scene context to effectively infer the missing information for these scenarios of high uncertainty.

\begin{figure}[t] 
  \centering 
  \includegraphics[width=\columnwidth]{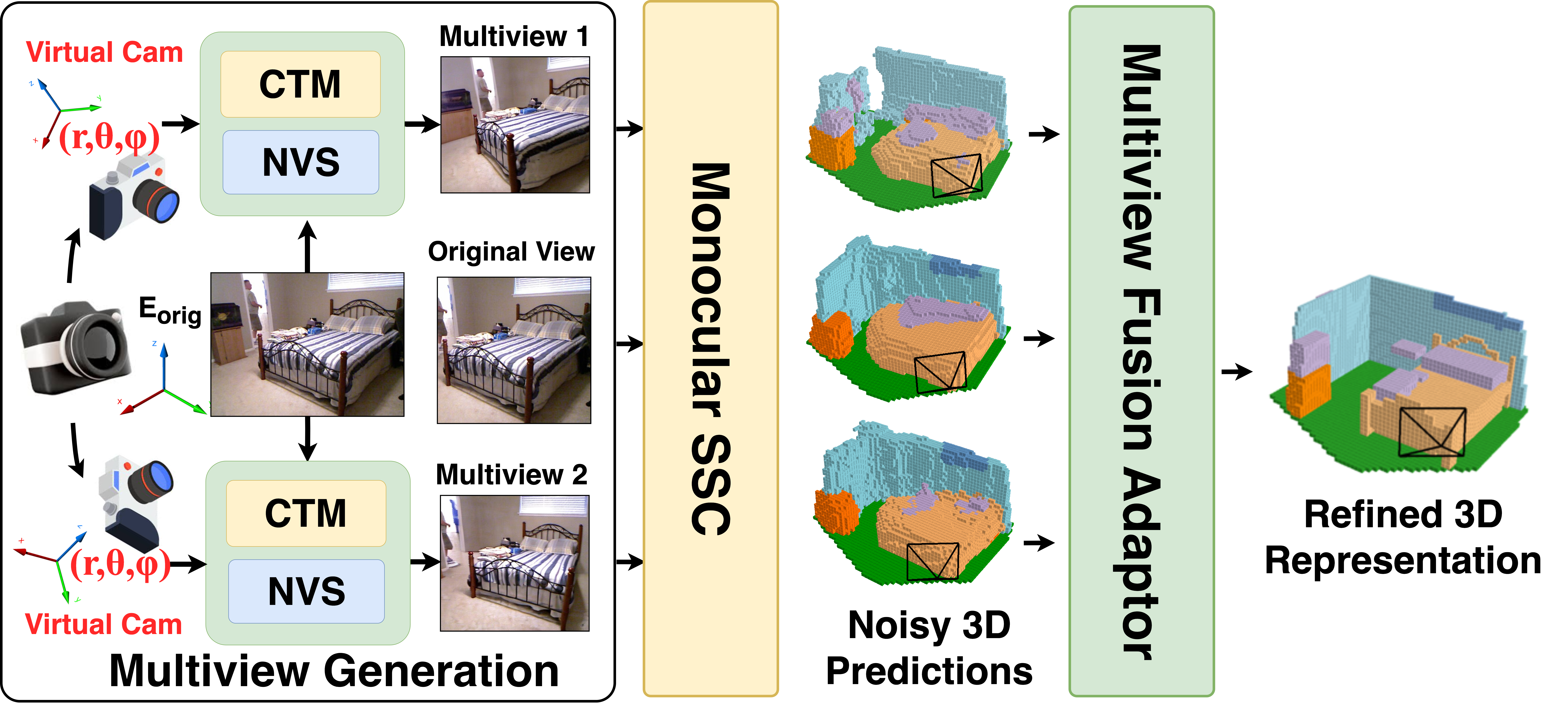} 
  \caption{High-level diagram of the proposed GenFuSE system. Novel View Synthesis techniques are used to generate context-relevant views that enhance predictions and reduce uncertainty of predictions of occluded regions.}
  \label{intro:concept}
\end{figure}

Outdoor SSC research has benefited from the rise in popularity of AV systems, with growing number of publicly available outdoor datasets and accompanying open challenges to encourage innovation in this space \cite{Caesar2020}. Despite apparent similarities, indoor SSC faces distinct challenges compared to outdoor SSC, notably a wider range of object size and shape aspect ratios \cite{Yu2024}. Therefore, innovations in outdoor SSC do not directly translate to indoor scenarios, hindering the adoption of those techniques for comparable improvements. Further, the richness in sensor information for outdoor applications differ from practical applications of indoor SSC. In practical applications, indoor navigation robots typically do not utilize the high-grade LiDAR or radar systems found in AVs. Consequently, SSC methods that rely solely on single RGB images are highly desirable, as they minimize the need for expensive depth sensors.

Monocular indoor SSC literature presents various techniques for reconstructing 3D semantic representations from single RGB images, including Feature Line of Sight Projection (FLoSP), neural network-based monocular depth estimation, and multitask learning \cite{Yu2024} \cite{Yao2023} \cite{Cao2022} \cite{JieLi2020}. However, the inherent lack of 3D geometric and spatial information in a single 2D image introduces depth, scale, and shape ambiguities, significantly hindering accurate SSC.

\begin{figure}[t] 
  \centering 
  \includegraphics[width=\columnwidth]{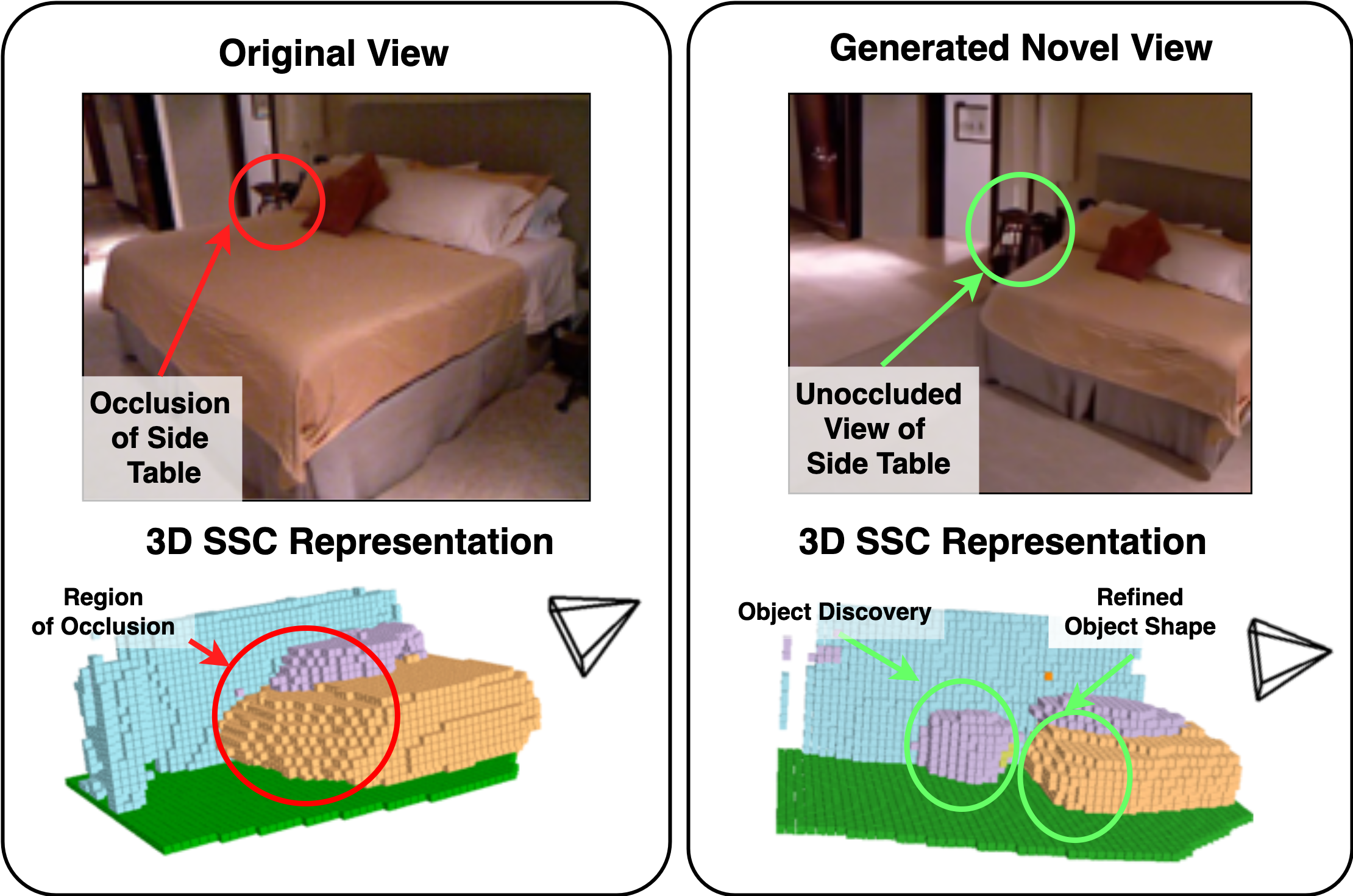} 
  \caption{Impact of occlusion on SSC from a single view (left) and the mitigation through novel view synthesis (right). Novel views provide line-of-sight access to occluded regions, enhancing object discovery and refining 3D shape representations.}
  \label{intro:nvs_example}
\end{figure}

To improve monocular indoor SSC, we propose the use of virtual multiviews, integrating novel view synthesis (NVS) for a one-shot multiview approach that reduces occlusion ambiguities. A practical example of this is shown in Figure \ref{intro:nvs_example}. While employing synthesized inputs for prediction enhancement is an emerging area \cite{bar2024} \cite{Janner2022} \cite{Li2023}, our work is the first to apply this to SSC. We present GenFuSE: a Virtual Multiview Generation and Fusion Framework for Semantic Scene Completion. Our main contributions are as follows:

\begin{enumerate}
    
    \item We propose a scene-constrained virtual camera placement strategy that optimizes the field of view for all virtual cameras
    \item We introduce a Multiview Fusion Adapter (MVFA) that enhances 3D prediction accuracy by fusing information from multiple views, effectively preserving spatial relationships and minimizing prediction uncertainty.
    \item Finally, we identify a Novelty-Consistency tradeoff, which provides crucial insight into the relationship between view synthesis and scene completion, highlighting the inherent balance between introducing new information and maintaining scene coherence.
\end{enumerate}

Results show IoU score improvements of up to 2.8\% of Scene Completion (SC) and 4.9\% for Semantic Scene Completion (SSC), demonstrating that GenFuSE can be utilized as a highly flexible virtual multiview generation and fusion framework to improve overall performance of SSC.

\section{Background}

\subsection{Related Works} SSC was introduced by Song et al \cite{Song2017}, where it was cited that scene completion and semantic labeling of depth maps are inherently highly coupled tasks. To that end, they introduced SSCNet, which converted input depth images into a 3D scene complete with occupancy and semantic labels per voxel. Following works in SSC introduced various approaches and techniques but ultimately relied on both visual and geometric inputs, such as depth information \cite{JieLi2020}. Monoscene \cite{Cao2022} was first of its kind to address indoor SSC relying on a single RGB image input- a challenging but common constraint for low-cost indoor robotic devices, by introducing a Features Line of Sight (FLoSP) module to transform the 2D image into a 3D scene without explicit geometric inputs. NDC-Scene \cite{Yao2023} tries to tackle feature and pose ambiguity inherent to \cite{Cao2022} by transforming the 2D image features first into a normalized 3D space called the Normalized Device Coordinate space (NDC), before the target world space. 3D convolutional networks are then used to learn spatial relationships and patterns directly in the 3D structure of the NDC space, which the authors show to result in improved performance over the previous methodology of operating directly in the target world space. Mostly recently, ISO \cite{Yu2024} showed that pretrained depth models can be leveraged in order to enhance the learning of 3D voxel features through a Dual Feature Line of Sight Projection (D-FLoSP) module and a multi-scale feature fusion strategy.

\subsection{Novel View Synthesis}

Novel view synthesis, the task of generating images of a scene from unseen viewpoints, has seen significant advancement throughout the last decade \cite{Muller2024}. While early techniques focused on geometry-based methods, and later, regressive models, they struggled to capture fine-grained details and generate high-resolution, photorealistic novel views \cite{Sajjadi2022}. This led to the exploration of generative models, particularly those based on implicit neural representations like Neural Radiance Fields (NeRFs) \cite{Mildenhall2020}. More recently, diffusion-based generative models have emerged as a powerful alternative for NVS. These methods, inspired by diffusion processes, learn to generate novel views by progressively denoising a random noise distribution conditioned on the input views. Diffusion models have shown impressive results, often exceeding the quality of the aforementioned techniques \cite{Seo2024}. NVS is particularly effective in indoor environments due to controlled lighting and detailed textures, enabling high-fidelity, realistic 2D scene reconstructions. 

\begin{figure*}[t] 
  \centering 
  \includegraphics[width=\textwidth]{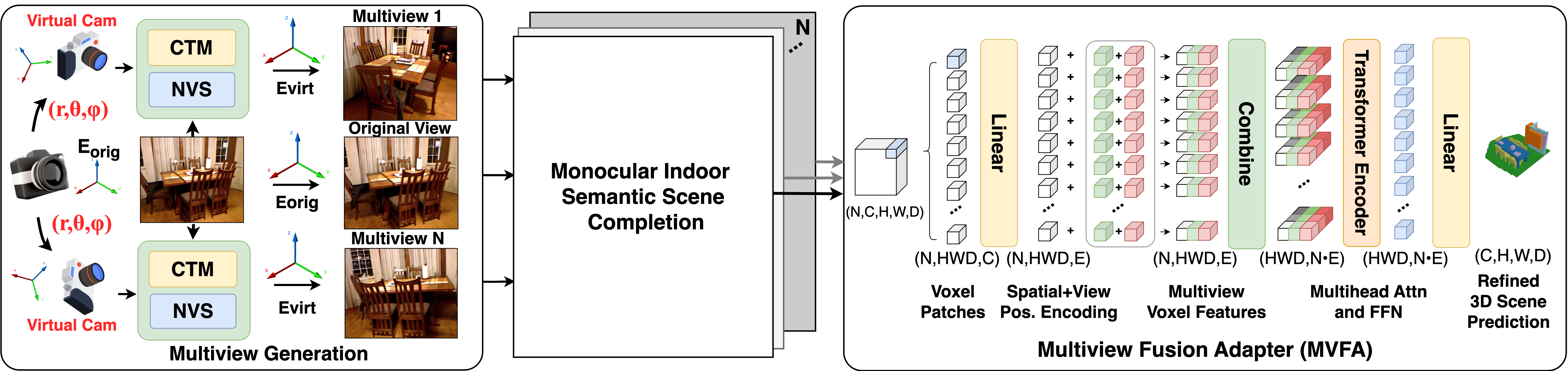} 
  \caption{System level diagram of GenFuSE. The Multiview Generation pipeline performs novel view synthesis to generate additional views that are passed into the SSC pipeline for prediction. These predictions are then passed into the Multiview Fusion Adaptor (MFVA) that fuse the multiview predictions into a refined 3D representation.}
  \label{genfuse:system_diagram}
\end{figure*}

\section{GenFuSE}

In this section, we outline the two major modules of the GenFuSE framework: (i) Virtual Multivew Scene Generation pipeline and (ii) Multiview Fusion Adaptor. A system-level diagram of GenFuSE shown in Figure \ref{genfuse:system_diagram}.

\subsection{Virtual Multiview Scene Generation Pipeline} 

To avoid the need for multiple physical cameras, our pipeline leverages virtual cameras placed at different viewpoints. Spherical transformations around the scene center are used to constrain virtual camera placement, guaranteeing that all generated views keep the scene within the field of view of the camera. This is achieved through the Coordinate Transformation Module (CTM) in conjunction with NVS. That is, the CTM translates standard rotations $(r, \theta, \phi)$, into camera poses suitable for both NVS and SSC modules:

\begin{equation}
\text{CTM}(\mathbf{r}, \theta, \psi) \rightarrow (\mathbf{E}_{\text{rel}}, \mathbf{E}_{\text{virt}})
\end{equation}

where $\textbf{r}$ is the mean-depth of the scene, computed by a deep neural depth network from the input image \cite{Yang2024}. Specifically, the CTM first computes the relative pose, $E_{rel}$, by performing a spherical rotation of the original camera around the mean-depth centroid of the scene, followed by a local rotation to re-orient the camera towards the centroid. This relative pose is then directly input into the NVS module for novel view generation. Subsequently, the camera pose of the novel view, aligned to the world coordinate system of the original camera, is derived as follows:
  
\begin{equation}
\mathbf{E}_{\text{virt}} = \mathbf{E}_{\text{rel}} \cdot \mathbf{E}_{\text{orig}}
\end{equation}

By maintaining controlled transformations around a fixed scene reference, we ensure scene-centric transformations regardless of camera placement, guaranteeing robustness across diverse camera configurations and scale across all scenes types. 

The resulting scene image and $E_{virt}$ is then passed into the SSC model. We can represent the 3D class-probability distribution output ($C$ classes for scene dimension $H$x$W$x$D$) of the SSC module for any given view, $i$, to be \textbf{$P_i$}. The collection of these predictions are then passed into the MVFA for multiview fusion.

\subsection{Multiview Fusion Adaptor}

We propose a transformer-based architecture for fusing multiple predictions in 3D semantic scene completion. We can describe the role of the MVFA is to intelligently fuse the predictions of multiple SSC 3D representation predictions into a single refined 3D representation. 
Given our aforementioned definition of $P_{i}$, we define the MVFA as follows:

\begin{equation}
\begin{aligned}
    \text{MVFA}: & \; \left\{ P_1, P_2, \dots, P_N \right\} \longrightarrow \textbf{P}, \\ 
    & \; \text{\textit{where}} \; P \in \mathbb{R}^{C \times H \times W \times D}
\end{aligned}
\end{equation}

This is achieved by first transforming the discrete channel probabilities $P_i$ into a higher-dimensional embedding space of dimension $E$ as:

\begin{align}
    X_i &= \text{Linear}(P_i), \quad X_i \in \mathbb{R}^{H \times W \times D \times E} \\
    X &= \text{Stack}(X_1, \dots, X_N), \quad X \in \mathbb{R}^{N \times H \times W \times D \times E}
\end{align}

To support the sequential formatting of the transformer, the 5 dimensional $(N, H, W, D, E)$ representation for each view is flattened into $H$x$W$x$D$ patches. Here, we introduce (i) spatial positional encoding (SPE) and (ii) view positional encoding (VPE):

\begin{equation}
\begin{aligned}
    SPE_{x, y, z} = \Big[ & \sin \left( \frac{x}{10000^{\frac{2k}{d}}} \right), \cos \left( \frac{x}{10000^{\frac{2k+1}{d}}} \right), \\
    & \sin \left( \frac{y}{10000^{\frac{2k}{d}}} \right), \cos \left( \frac{y}{10000^{\frac{2k+1}{d}}} \right), \\
    & \sin \left( \frac{z}{10000^{\frac{2k}{d}}} \right), \cos \left( \frac{z}{10000^{\frac{2k+1}{d}}} \right) \Big]
\end{aligned}
\end{equation}

\begin{equation}
\begin{aligned}
    VPE_{i} = \left[ \sin \left( \frac{i}{10000^{\frac{2k}{d}}} \right), \cos \left( \frac{i}{10000^{\frac{2k+1}{d}}} \right) \right]
\end{aligned}
\end{equation}


\begin{equation*}
\begin{aligned}
        \text{\textit{where}} \quad i &= 0, 1, 2, \dots, N-1, \quad x, y, z \in \mathbb{Z}, \\
        &k = 0, 1, 2, \dots, \frac{E}{2}-1
\end{aligned}
\end{equation*}




These encodings are directly added to the representation X as follows:

\begin{equation}
\begin{aligned}
X'_{i,h,w,d} &= X_{i,h,w,d} + \text{SPE}_{h, w, d} +  \text{VPE}_{i}
\end{aligned}
\end{equation}


The spatially and multi-view encoded embedding, $X'$, is reshaped into $(H \times W \times D)$ patches, each of dimension $N \cdot E$. These patches are then fed into a multi-head Transformer encoder, which utilizes the self-attention mechanism to fuse views, leveraging both spatial and multi-view context. Specifically, queries $Q$, keys $K$, and values $V$ are computed from the input $X'$ using learned weight matrices:

\begin{equation}
\begin{aligned}
Q = X' \mathbf{W_Q}, \quad K = X' \mathbf{W_K}, \quad V = X' \mathbf{W_V}
\end{aligned}
\end{equation}

where $\mathbf{W_Q}$, $\mathbf{W_K}$, $\mathbf{W_V}$ are learnable parameters. With $h$ attention heads and $d_{embed}$ as the embedding dimension, the multi-head attention output can then be expressed as:

\begin{equation}
\text{MultiHeadOutput} = \text{Concat}(\text{Attn}_1, \dots, \text{Attn}_h) \mathbf{W_O}
\end{equation}

where, 
\begin{equation}
\text{Attn}_i = \text{softmax}\left(\frac{Q_i K_i^T}{\sqrt{d_{\text{embed}} / h}}\right) V_i, \quad i = 1, 2, \dots, h
\end{equation}
and $\mathbf{W_o}$ is a learnable weight matrix used for projecting the concatenated multi-head attention outputs back to the original embedding space. Residual connection and layer normalization is performed on the multi-head output before passing it to feed forward network (FFN) and then additional residual connection and layer normalization:

\begin{align}
\text{Attn}_{\text{final}} &= \text{LayerNorm}\left(\text{MultiHeadOutput} + X'\right) \\
\text{FFNOutput} &= \text{Linear}(\text{Attn}_{\text{final}}) \\
\text{P} &= \text{LayerNorm}\left(\text{FFNOutput} + \text{Attn}_{\text{final}}\right)
\end{align}

\begin{center}
\textit{where,} \quad $\textbf{P} \in \mathbb{R}^{C \times H \times W \times D}$
\end{center}

By leveraging the attention mechanism of the transformer, the MVFA operates with a global receptive field, enabling the aggregation of context across both spatial and view dimensions for improved voxel prediction and reduced uncertainty.

\section{Experimentation}

This section describes the details of the experimental setup to producing the results reported, including dataset, implementation specifics, and network training.

\subsection{Dataset}

The NYUv2 dataset is a popular benchmark for indoor scene understanding, providing RGB-D images of various scenes in indoor environments \cite{Silberman}. The dataset also includes semantic labels for each pixel, across 13 semantic classes (empty, ceiling, floor, wall, window, chair, bed, sofa, table, TV, furniture, object). This dense semantic annotation enables the NYUv2 dataset to be well-suited for the semantic scene completion task. The NYUv2 dataset is divided into a training set and test set, containing 795 and 654 images, respectively, each of dimension 480x640.

\subsection{Implementations}

\subsubsection{Multiview Generation Pipeline}

The GenWarp \cite{Seo2024} model was chosen for the NVS component of our pipeline due to its practical advantages, including its diffusion architecture, readily available code, and pretrained checkpoints trained on indoor scenes of the NYUv2 dataset.

\subsubsection{Multiview Fusion Adaptor}

For our implementation of the MVFA, we transform voxel patches into a 24 dimensional embedding space through a linear layer. The transformer encoder has two attention heads and input and output embedding dimension of 24. A dropout rate of 0.1 was applied after both the multi-head attention and the feed-forward network (FFN) to prevent overfitting. These parameters were selected using hyperparameter tuning.

\begin{table*}[ht!]
\centering
\caption{Scene Completion (SC) and Semantic Scene Completion (SSC) performance metrics on the NYUv2 dataset \cite{Silberman}. IoU scores are provided. Higher scores represent a stronger prediction. Three SSC methods are used to show the flexibilty and benefit of GenFuSE. \textbf{*}Measurements on the baselines are obtained from retraining the network using available code.}
\label{tab:performance}
\begin{tabular}{lccccccccccccccc}
\toprule
\midrule
\multirow{2}{*}{\textbf{Method}} & \multirow{2}{*}{\textbf{N}} & \multirow{1}{*}{\textbf{SC}} & \multicolumn{13}{c}{\textbf{SSC}} \\ 
\cmidrule(lr){3-3} 
\cmidrule(lr){4-16}

 &  & IoU & Emp & Ceil & Floor & Wall & Window & Chair & Bed & Sofa & Table & TV & Furn & Obj & mIoU \\
\midrule
ISO\textbf{*} \cite{Yu2024} & 1 & 45.3 & 88.9 & 11.8 & 93.3 & 14.1 & 14.2 & 16.9 & 48.2 & 39.0 & 17.8 & 20.1 & 32.6 & 16.6 & 29.5  \\ %
 \textit{+ GenFuSE} & 3 & \textbf{46.3} & 90.0 & 9.8 & 92.8 & 13.8 & 13.7 & 17.6 & 48.6 & 39.7 & 17.6 & 24.7 & 33.4 & 17.2 & \textbf{29.9} \\ %
\midrule

&  & \textbf{+2.2\%} &  &  &  &  &  &  &  &  &  &  &  &  & \textbf{+1.4\%}  \\ %
\midrule
\midrule

NDCScene\textbf{*} \cite{Yao2023} & 1 & 42.5 & 88.7 & 11.3 & 93.4 & 11.8 & 13.7 & 15.4 & 48.5 & 36.3 & 16.8 & 21.9 & 29.2 & 12.3 & 28.2  \\ %
 \textit{+ GenFuSE} & 3 & \textbf{43.7} & 88.8 & 8.7 & 93.4 & 12.2 & 13.3 & 14.7 & 48.6 & 38.5 & 16.7 & 22.8 & 29.2 & 14.8 & \textbf{28.5} \\ %
\midrule

&  & \textbf{+2.8\%} &  &  &  &  &  &  &  &  &  &  &  &  & \textbf{+1.1\%}  \\ %
\midrule
\midrule

Monoscene\textbf{*} \cite{Cao2022} & 1 & 40.0 & 86.8 & 8.0 & 92.9 & 9.9 & 11.2 & 10.9 & 44.2 & 32.4 & 14.2 & 11.2 & 24.5 & 12.3 & 24.5  \\ %
 \textit{+ GenFuSE} & 3 & \textbf{41.0} & 87.8 & 7.9 & 93.5 & 10.8 & 10.7 & 12.7 & 46.6 & 34.7 & 13.3 & 15.7 & 24.5 & 12.9 & \textbf{25.7} \\ %
\midrule
&  & \textbf{+2.5\%} &  &  &  &  &  &  &  &  &  &  &  &  & \textbf{+4.9\%}  \\ %
\bottomrule
\end{tabular}
\end{table*}

\subsection{Training} 

NYUv2 images are passed directly to the SSC pipeline after being normalized for mean and standard deviation. Since the GenWarp module relies on inputs sized 512x512, the original image is center cropped and white-padded. While this results in some information loss at the edges of the scene, it is expected to be recovered through NVS. 

For training with GenFuSE, the Virtual Multiview Scene Generataion pipeline and MFVA are added to the baseline SSC architecture, and trained according to the schema of the SSC (all configurations and loss objective functions are preserved). Further, any pre-trained models in the original repository (i.e. EfficientNet-B7 in 2D UNet encoder, DepthAnythingV2) are kept frozen, along with the GenWarp module used for NVS. The remaining modules, including, but not limited to, the 3D decoder, depth refinement network (if applicable), and MVFA are trained with a learning rate of $10\mathrm{e}^{-4}$ and early stopping with 3 epoch patience tracking the mIoU. We integrate GenFuSE with ISO \cite{Yu2024}, NDCScene \cite{Yao2023}, and Monoscene \cite{Cao2022}. Baseline results were measured directly using the available code from the SSC model's respective official repository. 

\begin{table*}[ht!]
\centering
\caption{\textbf{Qualitative Analyses}: Outputs on NYUv2 for all baselines (ISO \cite{Yu2024}, NDCScene \cite{Yao2023} and Monoscene \cite{Cao2022}) and with GenFuSE framework. Three views are used (original view and 2 synthesized views, at $\pm 20^\circ$ azimuth rotations). GenFuSE results show refined object boundaries, smoother surfaces, and improved object completion across all baselines.}
\resizebox{\textwidth}{!}{
\begin{tabular}{c@{\hskip 2pt}|c@{\hskip 2pt}|cc|cc|cc} 
\toprule
\midrule
\multirow{2}{*}{\textbf{RGB}} & \multirow{2}{*}{\textbf{GT}} &
\multicolumn{2}{c|}{\textbf{ISO \cite{Yu2024}}} &
\multicolumn{2}{c|}{\textbf{NDCScene \cite{Yao2023}}} &
\multicolumn{2}{c}{\textbf{Monoscene \cite{Cao2022}}} \\

& & N = 1 & +GenFuSE (N=3) &
N = 1 & +GenFuSE (N=3\textbf{)} &
N = 1 & +GenFuSE (N=3) \\

\midrule

\includegraphics[width=0.11\textwidth]{figures/iso/NYU0947_0000_color.jpg} &
\includegraphics[width=0.11\textwidth]{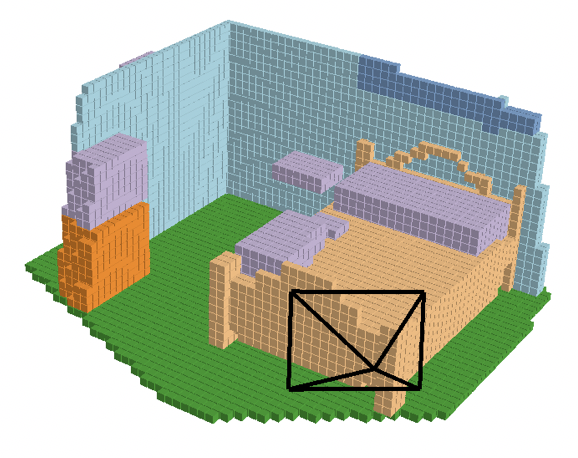} &
\includegraphics[width=0.11\textwidth]{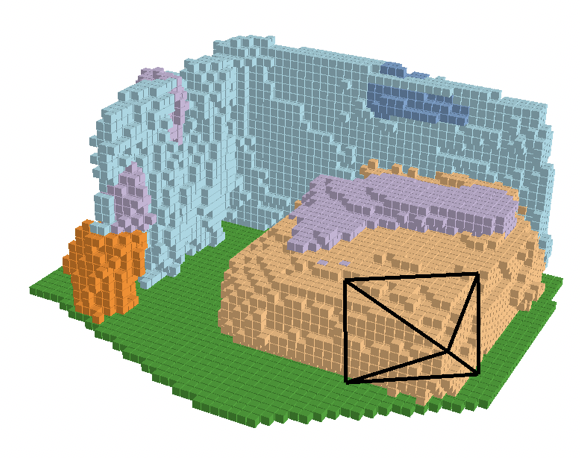} &
\includegraphics[width=0.11\textwidth]{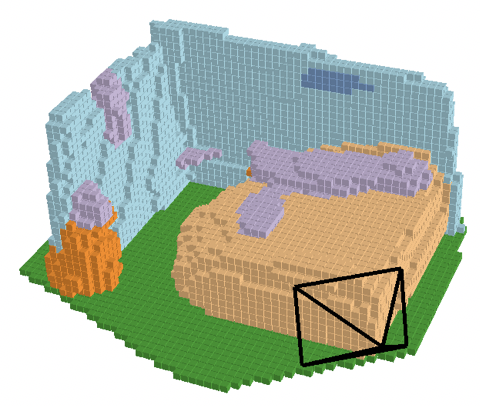} &
\includegraphics[width=0.11\textwidth]{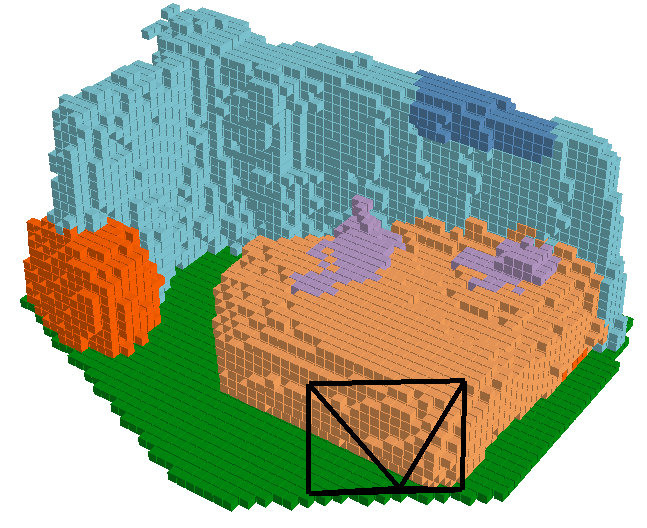} &
\includegraphics[width=0.11\textwidth]{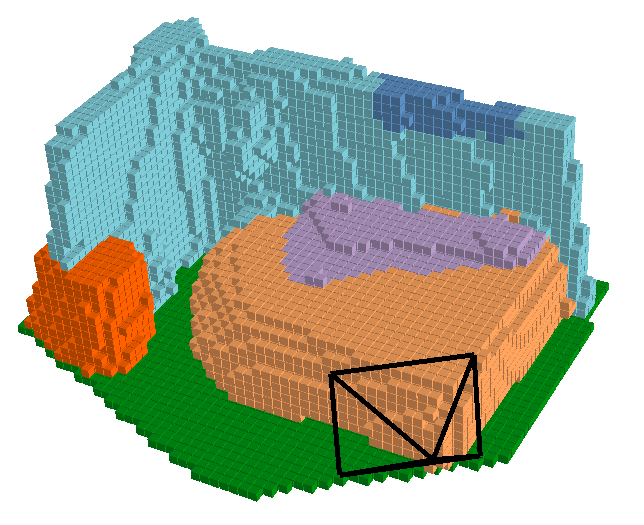} &
\includegraphics[width=0.11\textwidth]{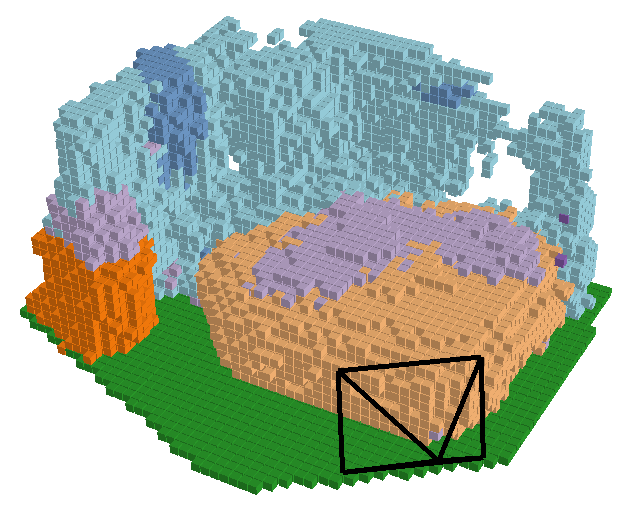} &
\includegraphics[width=0.11\textwidth]{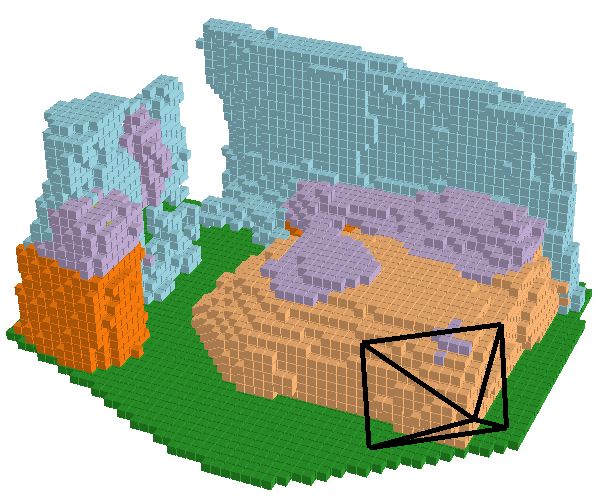} \\

\includegraphics[width=0.11\textwidth]{figures/iso/NYU0465_0000_color.jpg} &
\includegraphics[width=0.11\textwidth]{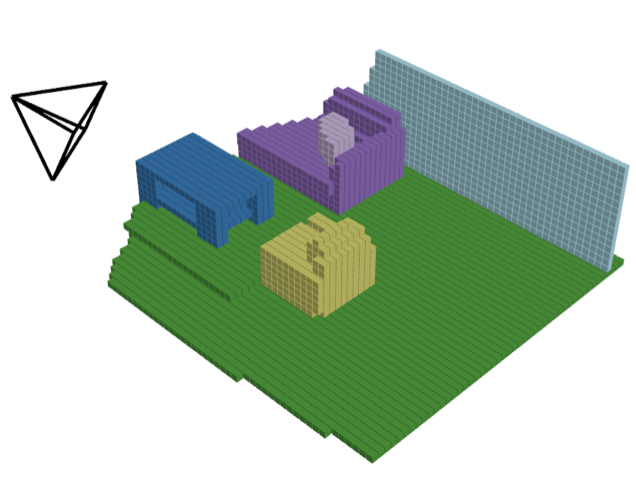} &
\includegraphics[width=0.11\textwidth]{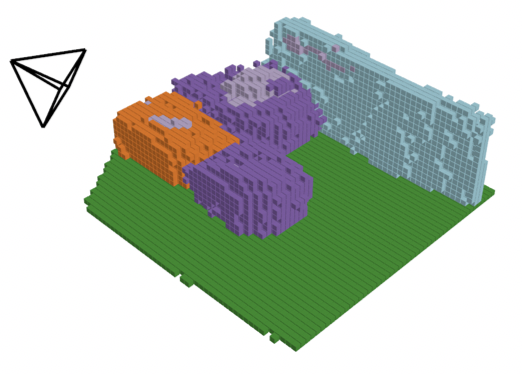} &
\includegraphics[width=0.11\textwidth]{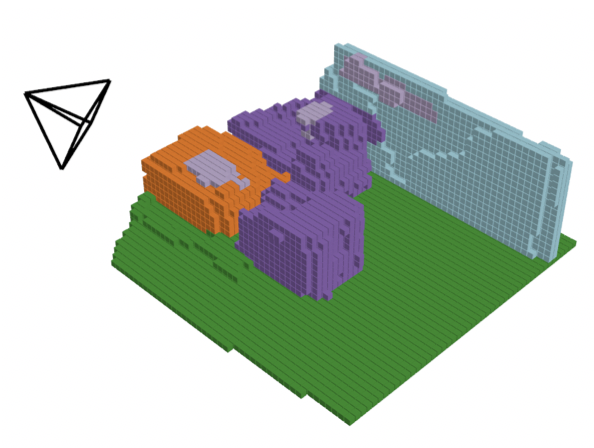} &
\includegraphics[width=0.11\textwidth]{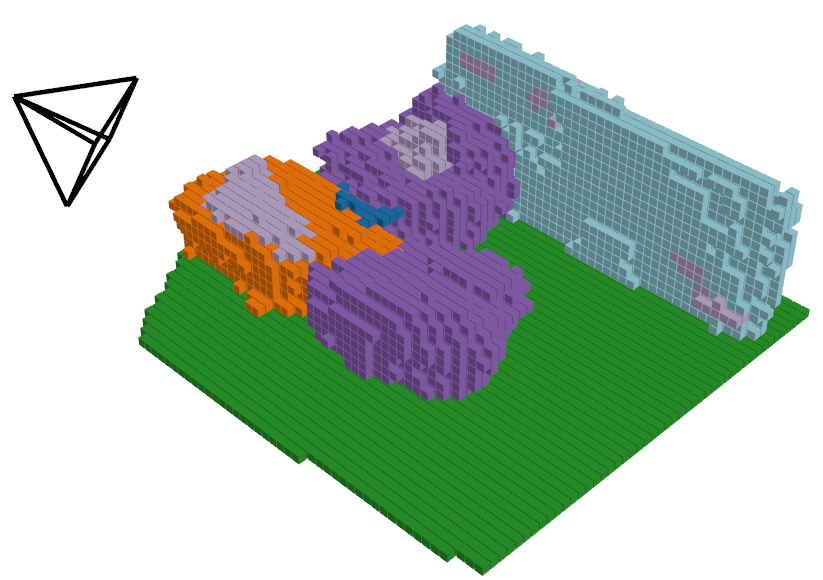} &
\includegraphics[width=0.11\textwidth]{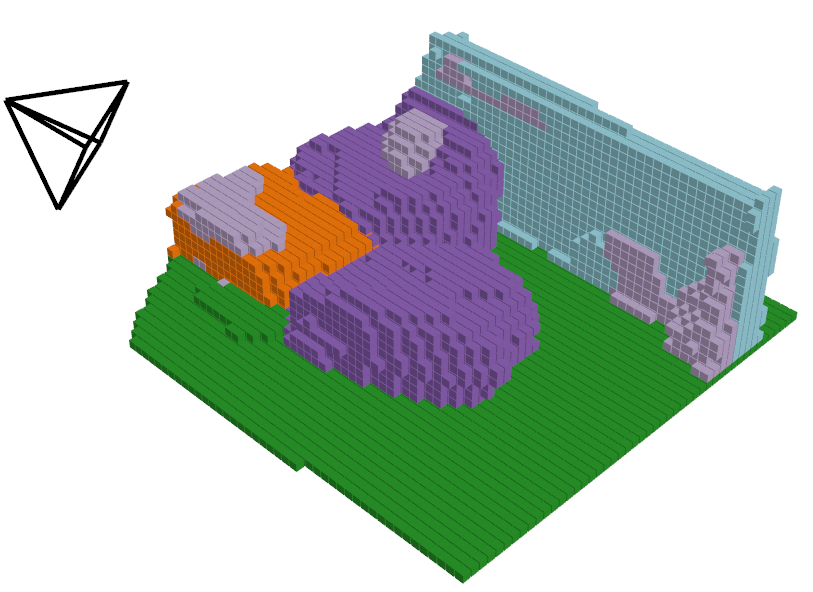} &
\includegraphics[width=0.11\textwidth]{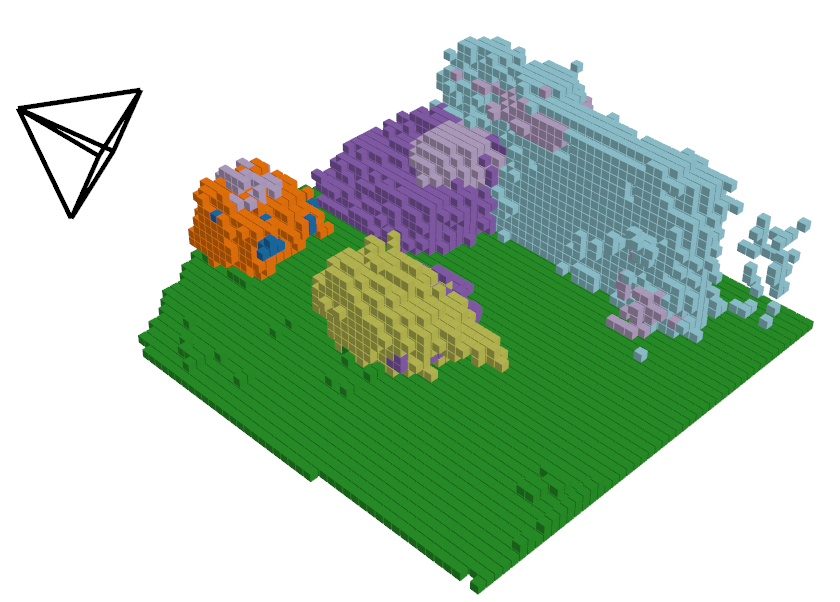} &
\includegraphics[width=0.11\textwidth]{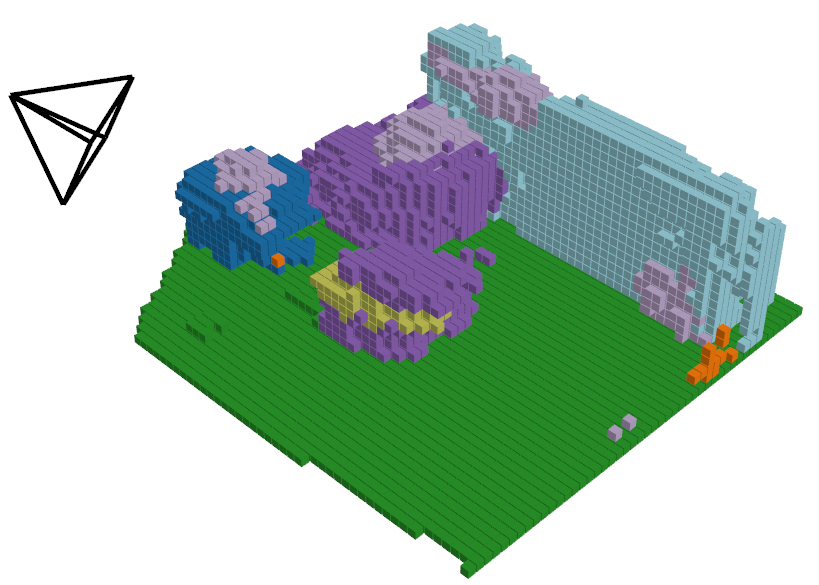} \\

\includegraphics[width=0.11\textwidth]{figures/iso/NYU1158_0000_color.jpg} &
\includegraphics[width=0.11\textwidth]{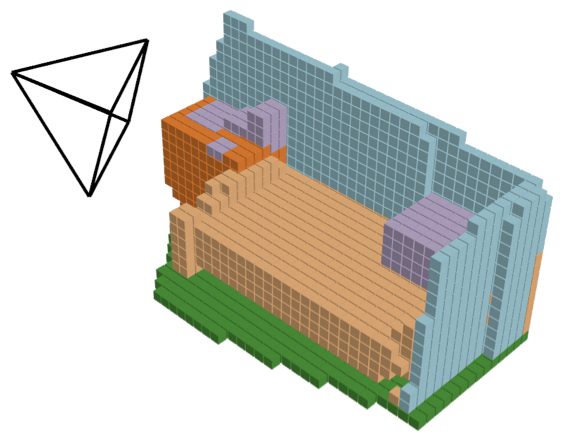} &
\includegraphics[width=0.11\textwidth]{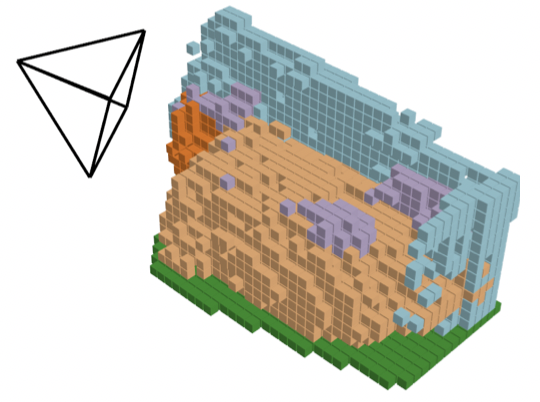} &
\includegraphics[width=0.11\textwidth]{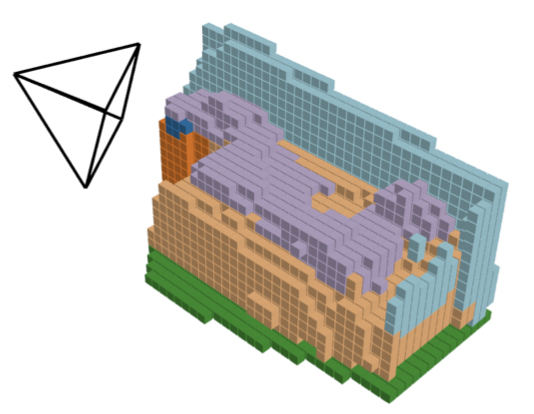} &
\includegraphics[width=0.11\textwidth]{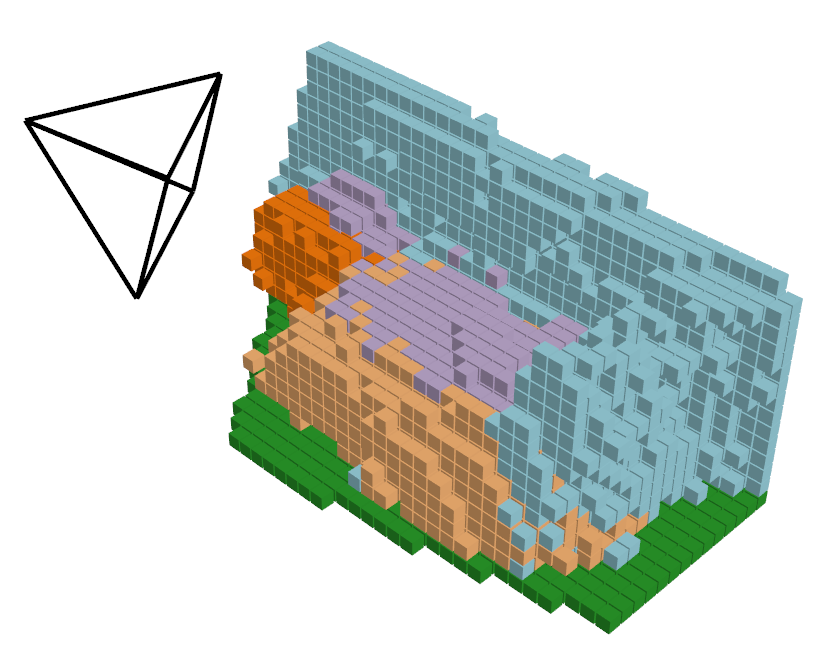} &
\includegraphics[width=0.11\textwidth]{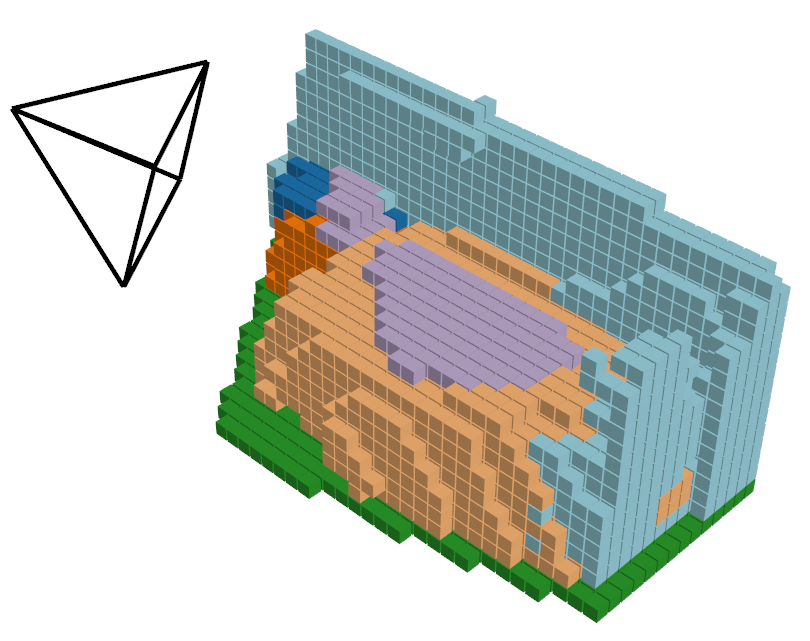} &
\includegraphics[width=0.11\textwidth]{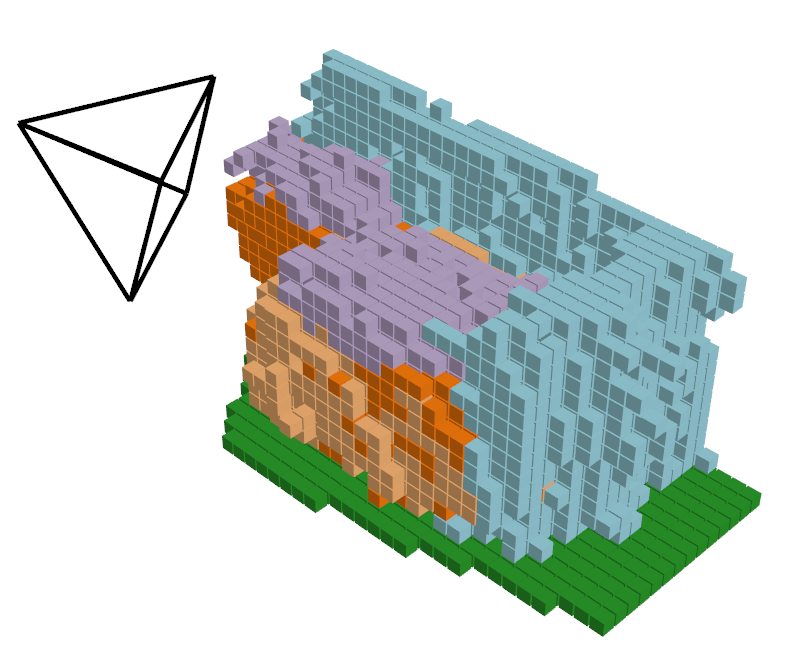} &
\includegraphics[width=0.11\textwidth]{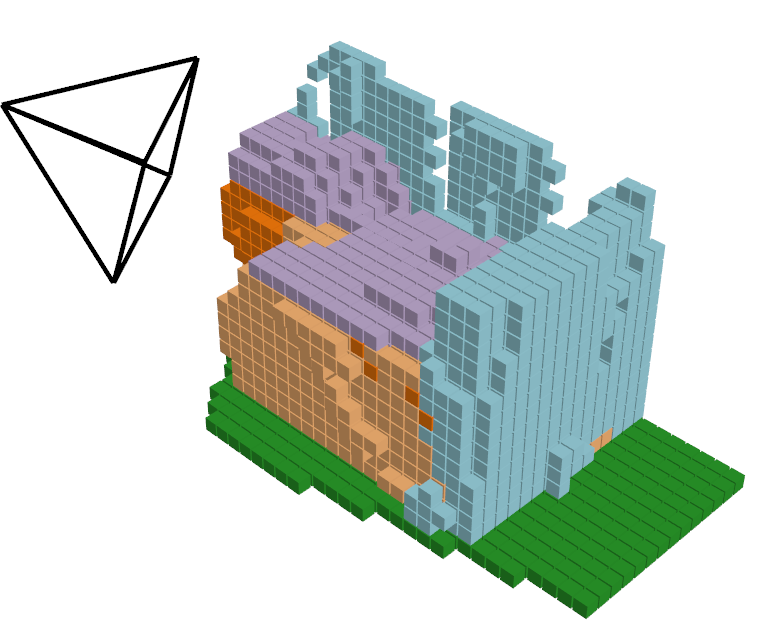} \\

\includegraphics[width=0.11\textwidth]{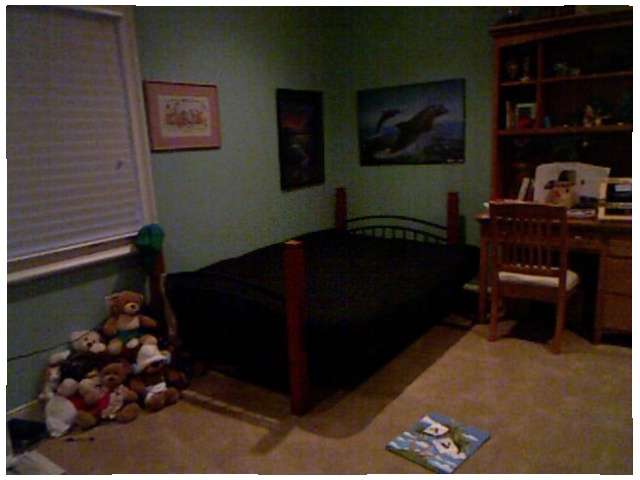} &
\includegraphics[width=0.11\textwidth]{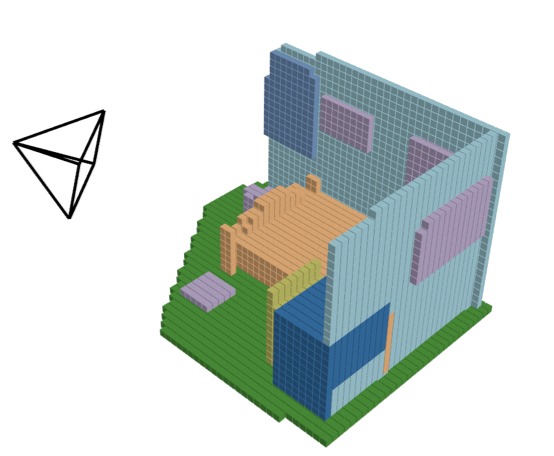} &
\includegraphics[width=0.11\textwidth]{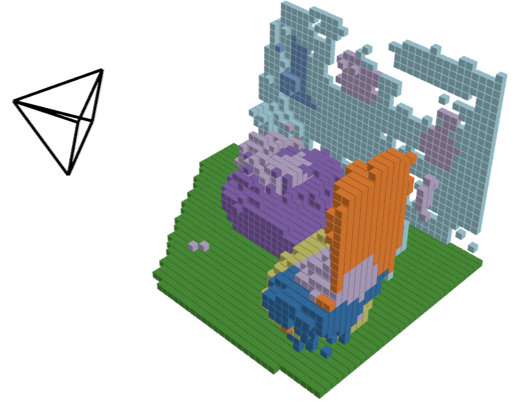} &
\includegraphics[width=0.11\textwidth]{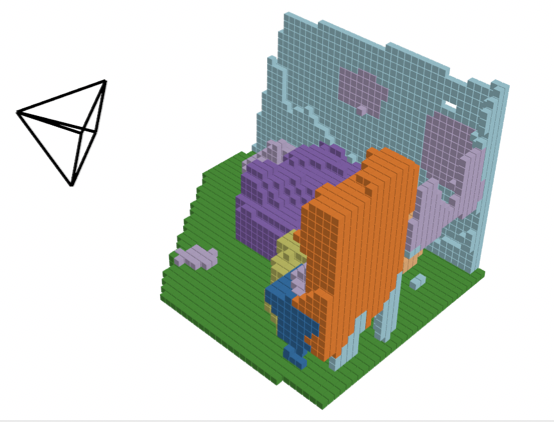} &
\includegraphics[width=0.11\textwidth]{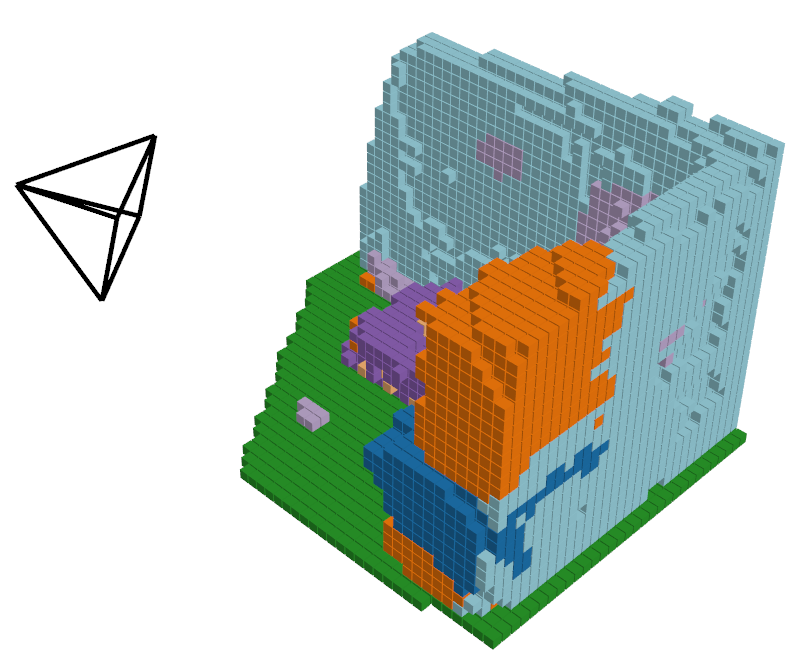} &
\includegraphics[width=0.11\textwidth]{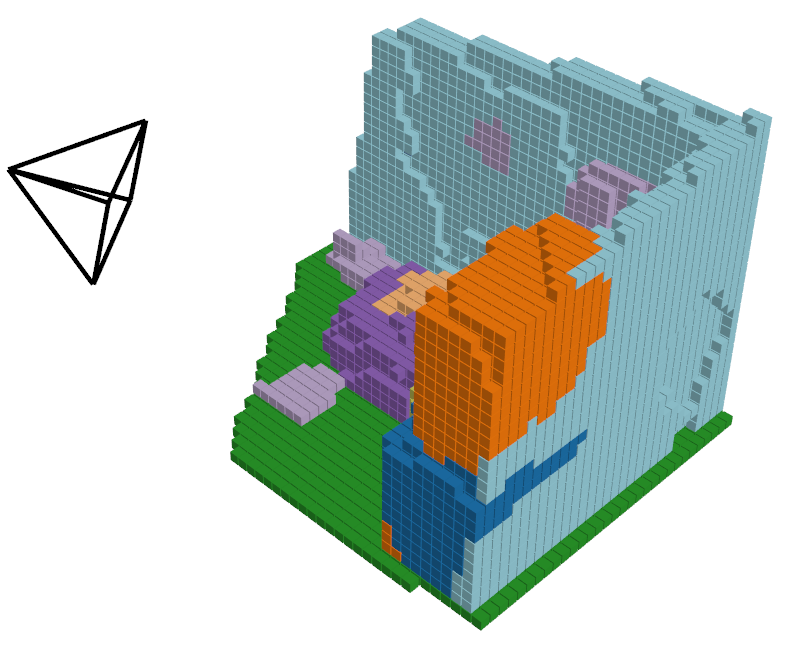} &
\includegraphics[width=0.11\textwidth]{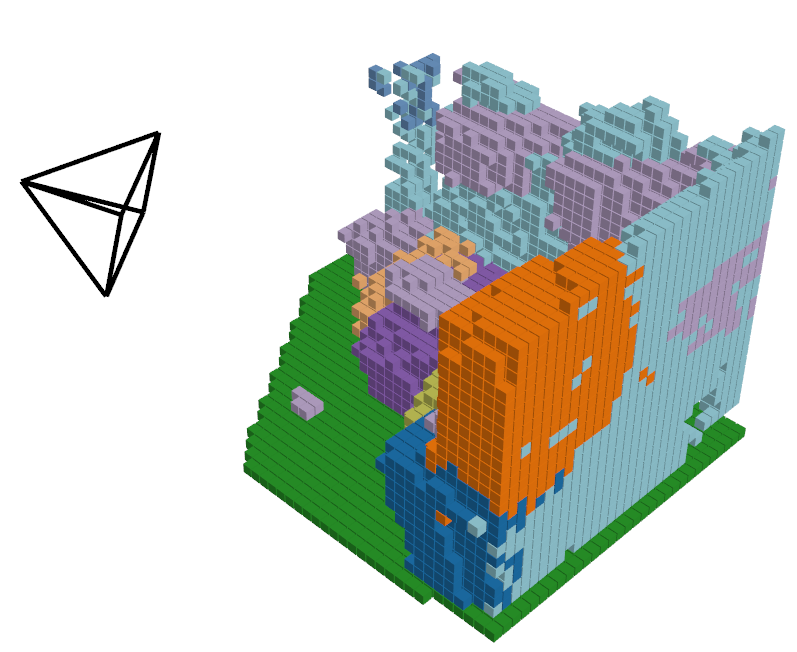} &
\includegraphics[width=0.11\textwidth]{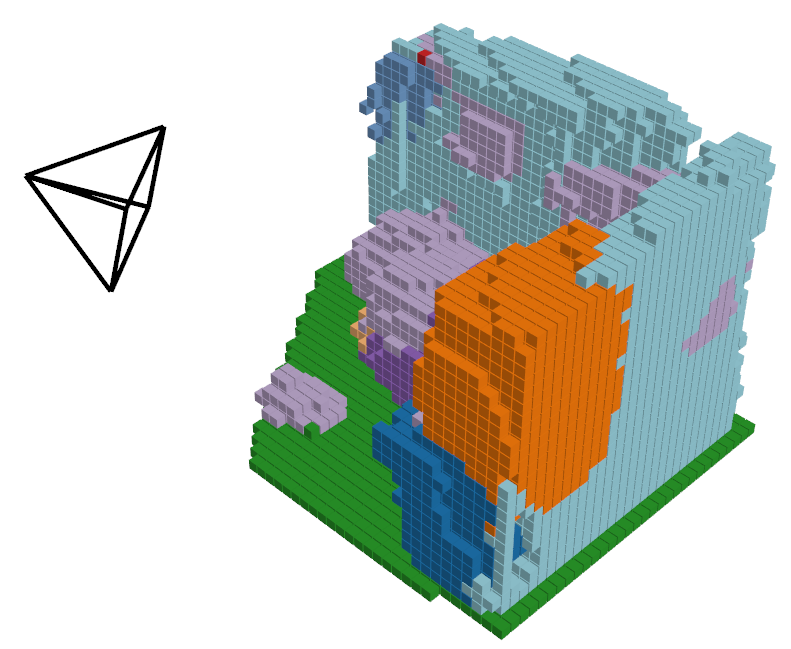} \\

\includegraphics[width=0.11\textwidth]{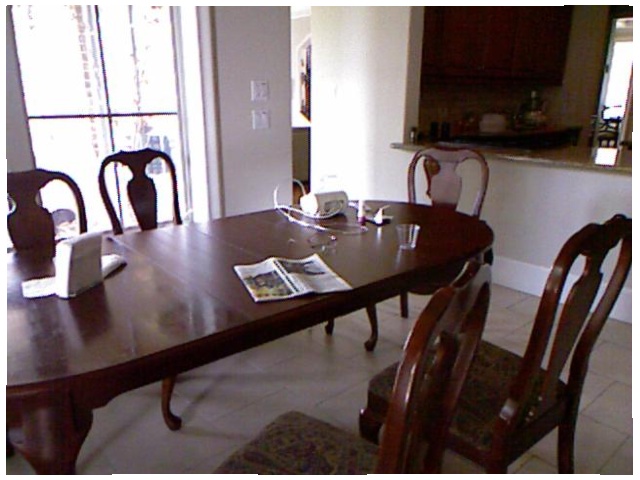} &
\includegraphics[width=0.11\textwidth]{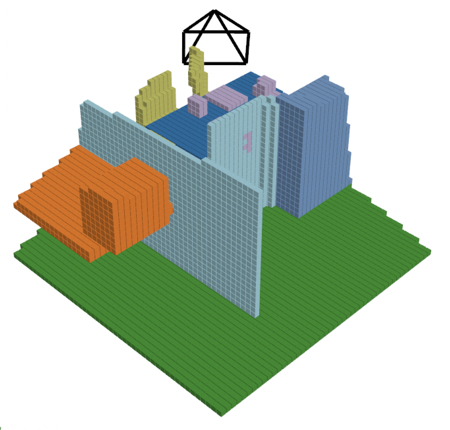} &
\includegraphics[width=0.11\textwidth]{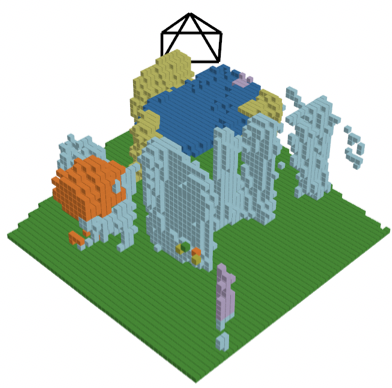} &
\includegraphics[width=0.11\textwidth]{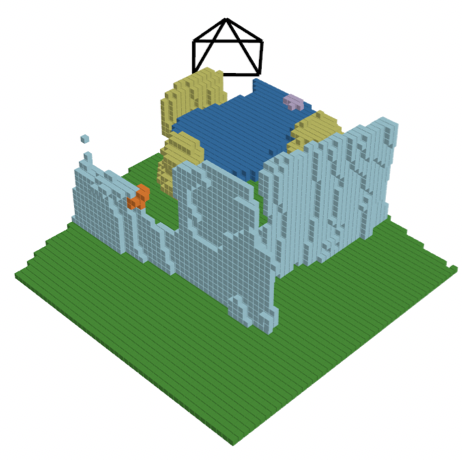} &
\includegraphics[width=0.11\textwidth]{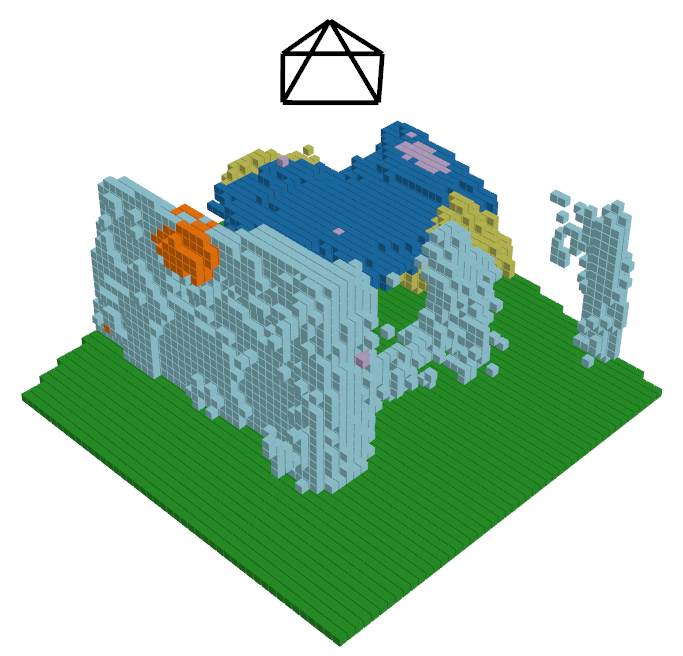} &
\includegraphics[width=0.11\textwidth]{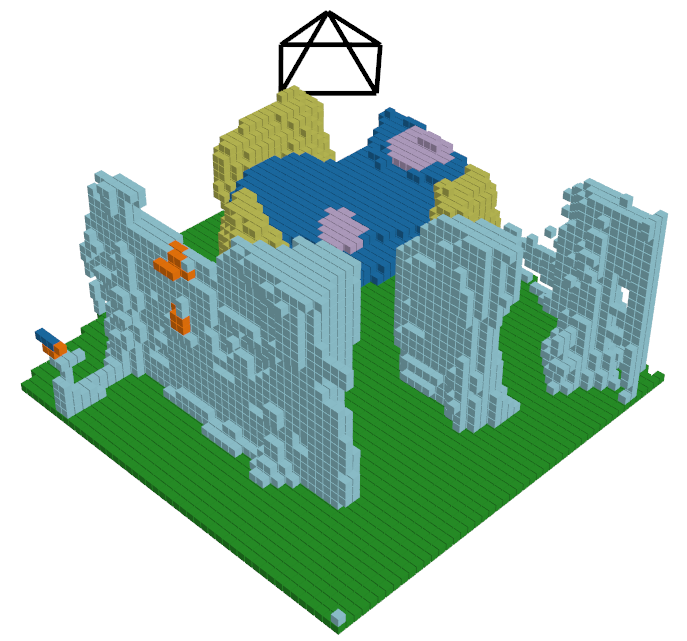} &
\includegraphics[width=0.11\textwidth]{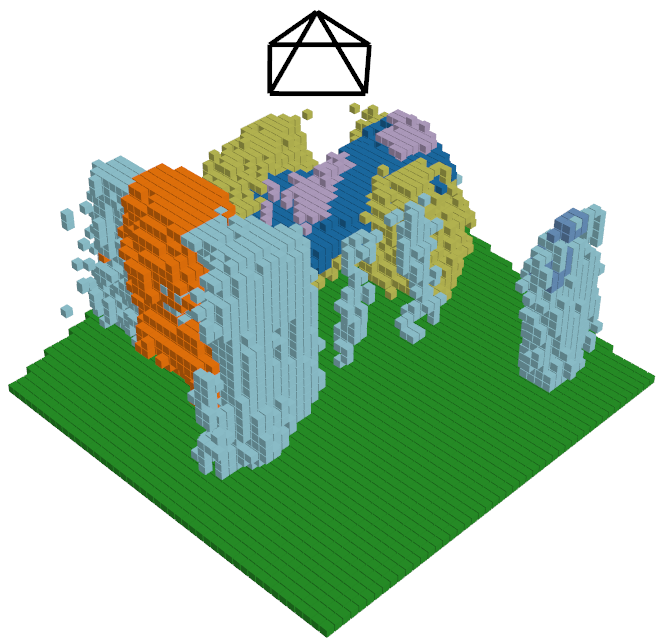} &
\includegraphics[width=0.11\textwidth]{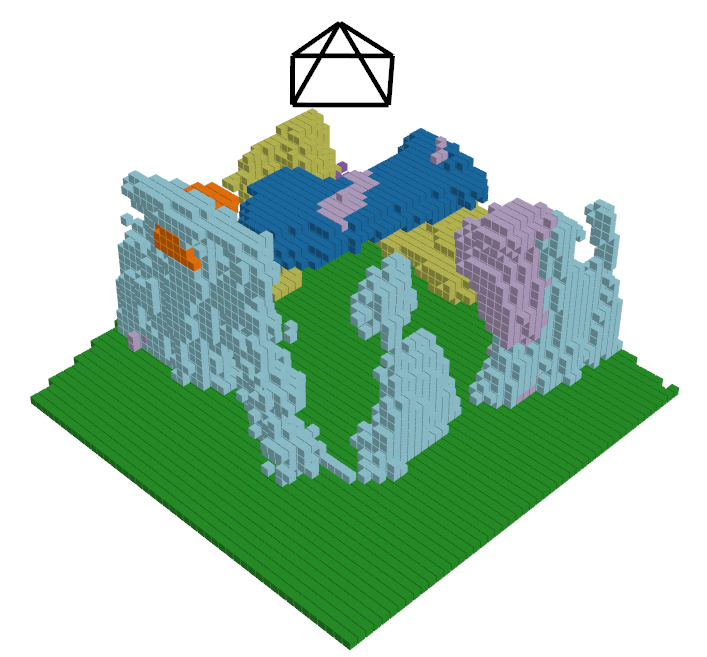} \\

\bottomrule
\end{tabular}\label{results:examples}
}
\end{table*}

\section{Results and Discussion}

We performed extensive experiments to study the benefit of the GenFuSE framework on mulitple SSC models, through qualitative and quantitative analyses. This section describes the results obtained along with a comparison to the baselines. 

\subsection{Quantitative Results}

Our results clearly demonstrate the flexibility and benefits of GenFuSE to complement existing methods for SSC. As shown in Table \ref{tab:performance}, we show that GenFuSE can enhance the overall quality of monocular SSC of up to 2.8\% for Scene Completion, and up to 4.9\% on Semantic Scene Completion (compared to reproduced results on three baselines: ISO \cite{Yu2024}, NDCScene \cite{Yao2023}, and Monoscene \cite{Cao2022}). 

The integration of additional views, through multiview fusion, partially mitigates feature scale, depth and shape ambiguity. Consequently, we observe improved object boundaries, leading to significant IoU gains, particularly for small to mid-size objects like tables, TVs, and furniture. However, we note a decrease in IoU for flat surfaces such as ceilings, walls, floors, and windows. This reduction is attributed to challenges in maintaining object consistency across novel views, as discussed further in Section \ref{res:ablation}. Additionally, the performance drop in ceiling and floor classes may be due to biases in the trained NVS model, which might prioritize center-scene content over edge-of-scene objects. This issue can be addressed in future work by retraining the NVS model to increase emphasis on these classes and by retraining the network to support the image dimensions of the NYUv2 dataset.

\subsection{Qualitative Results}\label{exp:qualitative}

Visual examples of SSC representations for the three baselines and our virtual multiview method are provided in Table \ref{results:examples}. We find that our approach yields two key enhancements: decreased noise and increased geometric consistency. Smoother object shapes, especially on large planar surfaces (Scenes 1-5), illustrate the reduction in noise. Scenes 1 and 3 demonstrate improved geometric consistency, with objects exhibiting more regular shapes and fewer distortions. Notably, single-view renderings often suffer from furniture distortions outside the camera's FoV, resulting in tapered side-surfaces. GenFuSE effectively addresses this, producing more uniform object shapes. Furthermore, GenFuSE models exhibit fewer wall gaps or holes (Scenes 4, 5), indicating improved completion.

By integrating multiple views, the MVFA model learns to discern geometric variations across different viewpoints, leading to a deeper understanding of 3D object structures and improved geometric consistency in the 3D occupancy map. Moreover, the global attention mechanism, enabled by the SPE and VPE, enhances 3D completion by leveraging contextual information from diverse 3D spatial locations to fill visual gaps.

\subsection{Ablation Study: Novelty-Consistency Tradeoff} \label{res:ablation}

\begin{table}[h]
\centering
\caption{ IoU scores are influenced by the level incremental viewing angle. Smaller incremental viewing angles mean less novel content needs to be generated compared to larger incremental viewing angles (azimuth rotation). The former also leads to less new information compared to the original scene, while the latter poses risks of hallucination.}
\label{tab:ablation} 
\begin{tabular}{lcccc}
\toprule
\midrule

\textbf{Method} & \textbf{\# Views} & \textbf{View $(\theta)$} & \textbf{IoU (SC)} & \textbf{mIoU (SSC)} \\
\midrule
\textbf{ISO*} \cite{Yu2024} & 1 & [$0^\circ$] & 45.3 & 29.5 \\
\textit{+ GenFuSE} & 3 & [$0^\circ$, $\pm$ $10^\circ$] & 44.8 & 29.3 \\
\textit{+ GenFuSE} & 3 & [$0^\circ$, $\pm$ $20^\circ$] & 46.3 & 29.9 \\
\textit{+ GenFuSE} & 3 & [$0^\circ$, $\pm$ $30^\circ$] & 42.3 & 27.2 \\
\bottomrule
\end{tabular}
\end{table}
While generative NVS models are capable of producing photo-realsitic, context-relevant novel scenes, they are also prone to hallucination. Hallucinations in NVS models can surface as artifacts such as blurring, ghosting, distortion, and new or missing objects \cite{Gonzalez2022}. The severity of hallucination can vary depending on the complexity of the scene, the quality of input images, and the amount of new content generated (i.e. related to the degree of overlap between the input view and the generated view). Since GenFuSE should be capable of performing in all indoor scenes (which inherently can vary in scene complexities) and input image quality is typically hardware-constrained, we instead investigate the effect of NVS hallucination on GenFuSE by varying the angular spread of the views (ranging from 0 to $\pm30^\circ$ in increments of $10^\circ$), and measure the corresponding SSC IoU scores for each configuration. Results are shown in Table \ref{tab:ablation}.

Based on the results of Table \ref{tab:ablation}, we can see a trade-off between scene consistency and novelty. We refer to consistency between scenes as having maintained coherence and stable properties across different views of the same scene (i.e. geometrically and semantically), and novelty as the introduction of new information (i.e. unseen regions, new object details). That is, with a single view, there is no novelty but high consistency. As additional views are incorporated, scene novelty increases, revealing more of the environment. With small incremental viewing angles $(\pm 10^{\circ})$, limited novelty results in slightly diminished scores. Increasing the viewing angle to $\pm 20^{\circ}$, however, significantly improves SC and SSC scores over the baseline. Conversely, generating content at excessively wide angles $(\pm 30^{\circ})$ leads to hallucination, reducing scene consistency. This dynamic, which we define as the Novelty-Consistency Tradeoff, is optimally balanced at $\pm 20^{\circ}$ angles for our multiview setup.

\section{CONCLUSIONS}

We explored the enhancement of monocular indoor scene completion by leveraging generative novel view synthesis methods. Our framework, GenFuSE, introduces virtual multiview cameras to generate additional viewpoints, enriching 3D representations. A Multiview Fusion Adaptor (MVFA), incorporating spatial and view positional encodings, effectively fuses predictions by leveraging global contextual information. We also identified the Novelty-Consistency Tradeoff, a limitation of novel view synthesis in Semantic Scene Completion (SSC) that highlights the challenge of balancing the introduction of new information with the preservation of scene structure. Demonstrating performance gains in both SC and SSC, GenFuSE establishes a precedent for leveraging synthesized inputs and is expected to serve as a foundational framework in the advancing fields of generative AI and monocular SSC.




\newpage

\bibliographystyle{ieeetr} 

\bibliography{references} 

\end{document}